\documentclass[journal]{IEEEtran} 
\usepackage{cite}

\usepackage{amsmath}
\usepackage{amssymb}
\usepackage{graphicx}
\usepackage{amsfonts}
\usepackage{algorithm}
\usepackage{algorithmic}
\usepackage{accents}
\usepackage{multirow}

\newcommand{\mat}[1]{\mathbf{#1}}

\ifCLASSINFOpdf
\else
\fi

\begin{document}
%
\title{Semi-Supervised Classification Through \\ the Bag-of-Paths Group Betweenness}
%
%
%

\author{Bertrand~Lebichot,
				Ilkka Kivim\"aki,
        Kevin~Fran\c{c}oisse
        \& Marco~Saerens,~\IEEEmembership{Member,~IEEE}
\thanks{The authors are with Universite Catholique de Louvain, ICTEAM \& LSM (e-mail: bertrand.lebichot@uclouvain.be).}
\thanks{This work was partially supported by the Elis-IT project funded by the \textquotedblleft R\'{e}gion wallonne\textquotedblright. We thank this institution for giving us the opportunity to conduct both fundamental and applied research.}
}

%
%

\markboth{Manuscript submitted for publication and subject to change}%
{Shell \MakeLowercase{\textit{et al.}}: Semi-Supervised Classification Through \\ the Bag of Paths Group Betweenness}
%



\maketitle

\begin{abstract}

This paper introduces a novel, well-founded, betweenness measure, called the Bag-of-Paths (BoP) betweenness, as well as its extension, the BoP group betweenness, to tackle semi-supervised classification problems on weighted directed graphs. The objective of semi-supervised classification is to assign a label to unlabeled nodes using the whole topology of the graph and the labeled nodes at our disposal. The BoP betweenness relies on a bag-of-paths framework \cite{KEVIN} assigning a Boltzmann distribution on the set of all possible paths through the network such that long (high-cost) paths have a low probability of being picked from the bag, while short (low-cost) paths have a high probability of being picked. Within that context, the BoP betweenness of node $j$ is defined as the sum of the a posteriori probabilities that node $j$ lies in-between two arbitrary nodes $i$, $k$, when picking a path starting in $i$ and ending in $k$. Intuitively, a node typically receives a high betweenness if it has a large probability of appearing on paths connecting two arbitrary nodes of the network. This quantity can be computed in closed form by inverting a $n \times n$ matrix where $n$ is the number of nodes. For the group betweenness, the paths are constrained to start and end in nodes within the same class, therefore defining a group betweenness for each class. Unlabeled nodes are then classified according to the class showing the highest group betweenness. Experiments on various real-world data sets show that BoP group betweenness outperforms all the tested state-of-the-art methods \cite{Callut-2007,Pan-2004,Zhou-2003,Zhu-2003}. The benefit of the BoP betweenness is particularly noticeable when only a few labeled nodes are available.
\end{abstract}

\begin{IEEEkeywords}
Graph and network analysis, network data, graph mining, betweenness centrality, kernels on graphs, semi-supervised classification.
\end{IEEEkeywords}

%
\IEEEpeerreviewmaketitle

\section{Introduction}
%
%
%
%
\IEEEPARstart{A}{s} is well-known, the goal of a classification task is to automatically assign data to predefined classes. Traditional pattern recognition, machine learning or data mining classification methods require large amounts of labeled training instances, which are often difficult to obtain.
The effort required to label the data can be reduced using, for example, semi-supervised learning methods. This name comes from the fact that the data used is a mixture of data used for supervised and unsupervised learning (see, e.g., \cite{Zhu-2008,Zhu-2009} for a comprehensive introduction). Actually, semi-supervised learning methods learn from both labeled and unlabeled instances. This allows to reduce the amount of labeled instances needed to achieve the same level of classification accuracy.

Graph-based semi-supervised classification has received a growing focus in recent years. The problem can be described as follows: given an input graph with some nodes labeled, the problem is to predict the missing node labels. This problem has numerous applications such as classification of individuals in social networks, linked documents (e.g. patents or scientific papers) categorization, or protein function prediction, to name a few.
In this kind of application (as in many others), unlabeled data are usually available in large quantities and are easy to collect: friendship links can be recorded on Facebook, text documents can be crawled from the internet and DNA sequences of proteins are readily available from gene databases. Given a relatively small labeled data set and a large unlabeled data set, semi-supervised algorithms can infer useful information from both sources. 

Still another way to reduce the effort required to label the training data is to use an active learning framework. Active learning methods reduce the number of labeled data required for learning by intelligently choosing which instance to ask to be labeled next (see, e.g., \cite{Settles-2012}). However, this second approach will not be studied in this paper and is left for future work.

This paper tackles this problem within the \textbf{bag-of-paths} (BoP) \textbf{framework} \cite{KEVIN} capturing the global structure of the graph with, as building block, network paths. More precisely, we assume a weighted directed graph or network $G$ where a cost is associated to each arc. We further consider a bag containing all the possible paths (also called walks) between pairs of nodes in $G$. Then, a Boltzmann distribution, depending on a temperature parameter $T$, is defined on the set of paths such that long (high-cost) paths have a low probability of being picked from the bag, while short (low-cost) paths have a high probability of being picked. In this probabilistic framework, the \textbf{BoP probabilities}, $\text{P}(s=i,e=j)$, of sampling a path starting in node $i$ and ending in node $j$ can easily be computed in closed form by a simple $n \times n$ matrix inversion where $n$ is the number of nodes.

Within this context, a betweenness measure quantifying to which extent a node $j$ is in between two nodes $i$ and $k$ is defined. More precisely, the \textbf{BoP betweenness} of a node $j$ of interest is defined quite naturally as the sum of the a posteriori probabilities that node $j$ (intermediate node) lies in between two arbitrary nodes $i$, $k$, $\text{bet}_{j} = \sum_{i=1}^{n} \sum_{k=1}^{n} \text{P}(int=j|s=i,e=k)$, when picking a path starting in $i$ and ending in $k$. Intuitively, a node receives a high betweenness if it has a large probability of appearing on paths connecting two arbitrary nodes of the network.

For the \textbf{group betweenness}, the paths are constrained to start and end in nodes of the same class, therefore defining a group betweenness between classes, $\text{gbet}_{j}(\mathcal{C}_i,\mathcal{C}_k) = \text{P}(int=j|s \in \mathcal{C}_i,e \in \mathcal{C}_k)$. Unlabeled nodes are then classified according to the class showing the highest group betweenness when starting and ending within the same class.

In summary, this work has three main contributions:

\begin{itemize}
		\item It develops both a betweenness measure and a group betweenness measure from a well-founded theoretical framework, the bag-of-paths framework introduced in \cite{KEVIN}. These two measures can be easily computed in closed form.
		\item This group betweenness measure provides a new algorithm for graph-based semi-supervised classification.
		\item It assesses the accuracy of the proposed algorithm on thirteen standard data sets and compares it to state-of-the-art techniques. The obtained performances are highly competitive in comparison with the other graph-based semi-supervised techniques.
	\end{itemize}

In this paper, the BoP classifier (or just BoP) will refer to the semi-supervised classification algorithm based on the bag-of-paths group betweenness, which is developed in Section \ref{BoPClass}.

The paper is organized as follows. Section \ref{BgNot} introduces background and notations, mainly the bag-of-paths and the bag-of-hitting-paths models. Then, related works in semi-supervised classification is discussed in Section \ref{RelW}. The bag-of-paths betweenness and group betweenness centralities are introduced in Section \ref{BoP}. This enables us to derive the BoP classifier in Section \ref{BoPClass}. Then experiments involving the BoP classifier and classifiers discussed in the related works section will be performed in Section \ref{Exp}. Results and discussions of those experiments can be found in Section \ref{ResDis}. Finally, Section \ref{CCL} concludes this paper and opens a reflexion for further works.

\section{Background and notations}
\label{BgNot}

This section aims to introduce the theoretical background and notations used in this paper. First, graph-based semi-supervised classification will be discussed in Section \ref{Notations}, then the bag-of-paths model introduced in \cite{KEVIN} will be summarized in Section \ref{BoPK}. Finally, the bag-of-hitting-paths model will be introduced in Section \ref{BoHP}.


\subsection{Graph-based semi-supervised classification}
\label{Notations}

Consider a weighted directed graph or network, $G$, strongly connected  with a set of $n$ nodes $\mathcal{V}$ (or vertices) and a set of edges $\mathcal{E}$. Also consider a set of classes, $\mathcal{C}$, with the number of classes equals to $m$. Each node is assumed to belong to at most one class, since the class label can also be unknown. To represent the class memberships, an $n\times m$-dimensional indicator matrix, $\mathbf{Y}$, is used. On each of its rows, it contains as entries $1$ when the corresponding node belongs to class $c$ and $0$ otherwise ($m$ zeros on line $i$ if node $i$ is unlabeled). The $c$-th column of $\mathbf{Y}$ will be denoted $\mathbf{y}^{c}$. To each edge between node $i$ and $j$ is associated a positive number $c_{ij}>0$. This number represents the \textbf{immediate cost} of transition between node $i$ and $j$. If there is no link between $i$ and $j$, the cost is assumed to take a large value, denoted by $c_{ij} = \infty$. The \textbf{cost matrix} $\mathbf{C}$ is an $n\times n$ matrix containing the $c_{ij}$ as elements.

Moreover, a \textbf{natural random walk} on $G$ is defined in the standard way. In node $i$, the random walker chooses the next edge to follow according to reference transition probabilities
\begin{equation}
	p_{ij}^{\text{ref}}=\frac{1/c_{ij}}{{\displaystyle \sum_{j'=1}^{n}}(1/c_{ij'})}
	\label{Pref}
\end{equation}
representing the probability of jumping from node $i$ to node $j \in \mathcal{S}ucc(i)$, the set of successor nodes of $i$. The corresponding transition probabilities matrix will be denoted as $\mathbf{P}^{\text{ref}}$. In other words, the random walker chooses to follow an edge with a probability proportional to the inverse of the immediate cost (apart from the sum-to-one normalization), therefore favoring edges having a low cost.


\subsection{The bag-of-paths framework}
\label{BoPK}

The framework introduced in \cite{KEVIN} is extended in this paper in order to define new betweenness measures. The bag-of-paths (BoP) model can be considered as a motif-based model \cite{Milo-2002,Arenas-2008} using, as building blocks, paths of the network. In the next section, hitting paths will be used instead, as motifs. The BoP framework is based on the probability of picking a path $i \leadsto j$ starting at a node $i$ and ending in a node $j$ from a virtual bag containing all possible paths of the network. Let us define $\mathcal{P}_{ij}$ as the set of all possible paths connecting node $i$ to node $j$, including loops. Let us also define the set of all paths through the graph as $\mathcal{P}=\bigcup_{i,j=1}^{n}\mathcal{P}_{ij}$. Each path is weighted according to its \textbf{total cost} so that the likelihood of picking a low-cost path is higher that picking a high-cost path (low-cost paths are therefore favoured). The \textbf{total cost} of a path $\wp$, $\tilde{c}(\wp)$, is defined as the sum of the individual transition costs $c_{ij}$ along $\wp$. A path $\wp$ (also called a walk) is a sequence of transitions to adjacent nodes on $G$ (loops are allowed), initiated from a starting node $s$, and stopping in an ending node $e$.


The potentially infinite set of paths in the graph is enumerated and a probability distribution is assigned to each individual path: the longer (high-cost) the path, the smaller the probability of following it. This probability distribution depends on the \textbf{inverse-temperature parameter}, $\theta = \frac{1}{T} > 0$, controlling the exploration carried out in the graph. In \cite{KEVIN}, the authors assume that the probability of picking a path $\mathcal{P}$ from the bag follows a \textbf{Boltzmann distribution} (for details, see \cite{KEVIN}):
\begin{equation}
			\text{P}(\wp) = 
			\frac		{\tilde{\pi}^{\text{ref}}(\wp) \exp[-\theta \tilde{c}(\wp)] }  
							{{\displaystyle \sum\limits_{\wp' \in \mathcal{P}}}\tilde{\pi}^{\text{ref}}(\wp') \exp[-\theta \tilde{c}(\wp')] }
			\label{Boltzmann}
			\end{equation}
which is derived in \cite{KEVIN} from a cost minimization perspective subject to a relative entropy constraint. Recall that $\mathcal{P}$ is the set of all paths through the graph and $\tilde{\pi}^{\text{ref}}$ is the product of the transition probabilities $p_{ij}^{\text{ref}}$ along the path $\wp$. As expected, short (low-cost) paths are favored since they have a larger probability of being picked. Furthermore, when $\theta \rightarrow 0^{+}$, the path probabilities reduce to the probabilities given by the natural random walk on the graph. On the other hand, when $\theta$ becomes large, the probability distribution defined by  (\ref{Boltzmann}) is more and more biased towards shorter paths (the most likely paths are the shortest ones).

The \textbf{bag-of-paths probability} is the quantity $\text{P}(s = i , e = j)$. It is defined as the probability of drawing a path starting from node $i$ and ending in node $j$ from the bag-of-paths:
\begin{equation}
			\text{P}(s = i , e = j) = 
			\frac		{\displaystyle \sum\limits_{\wp \in \mathcal{P}_{ij}}\tilde{\pi}^{\text{ref}}(\wp) \exp[-\theta \tilde{c}(\wp)] }  
							{\displaystyle \sum\limits_{\wp' \in \mathcal{P}}\tilde{\pi}^{\text{ref}}(\wp') \exp[-\theta \tilde{c}(\wp')] }
			\label{BagOfP}
\end{equation}
where it is assumed for the reference probabilities that the starting and ending nodes are selected thanks to a uniform probability.
In \cite{KEVIN}, the authors have also shown that this probability can be easily calculated as
\begin{equation}
\text{P}(s=i,e=j) = \frac{z_{ij}}{\displaystyle \sum_{i'=1}^{n}\displaystyle \sum_{j'=1}^{n}z_{i'j'}} = \frac{z_{ij}}{\mathcal{Z}} , \text{ with } \mathbf{Z}=(\mathbf{I}-\mathbf{W}\mathbf{)}^{-1}
\label{Eq_bag_of_paths_probabilities01}
\end{equation}
where $\mathcal{Z} = \sum_{i=1}^{n} \sum_{j=1}^{n}z_{ij}$ is the \textbf{partition function }and $z_{ij}$ is the element $i,j$ of matrix $\textbf{Z}$. In (\ref{Eq_bag_of_paths_probabilities01}), matrix $\mathbf{Z}$ is called the fundamental matrix and is computed from the $n \times n$ matrix
\begin{equation}
	\textbf{W} = \textbf{P}^{\text{ref}} \circ \exp[-\theta  \mathbf{C}]
	\label{W}
\end{equation}
where $\circ$ is the elementwise (Hadamard) matrix product and the logarithm and exponential
functions are taken elementwise. The entries of \textbf{W} are therefore $w_{ij}=p_{ij}^{\text{ref}}\exp\left[-\theta  c_{ij}\right]$. Notice that $\text{P}(e = j | s = i)$ is not symmetric and that variables $z_{ij}$ are defined as \cite{KEVIN}
\begin{equation}
 z_{ij} = \sum_{\wp\in\mathcal{P}_{ij}} \tilde{\pi}^{\text{ref}}(\wp) \exp[-\theta \tilde{c}(\wp)] 
\end{equation}

We now turn to a variant of the bag-of-paths, the \textbf{bag-of-hitting-paths}.


\subsection{The bag-of-hitting-paths framework}
\label{BoHP}

The idea behind the bag-of-hitting-paths model is the same as the bag-of-paths model but the set of paths is now restricted to trajectories in which the ending node does not appear more than once, i.e. it only appears at the end of the path. In other words, no intermediate node on the path is allowed to be the ending node $j$ (node $j$ is made absorbing) and the motifs are now the hitting paths. Hitting paths will play an important role in the derivation of the BoP betweenness. In that case, it can be shown \cite{KEVIN} that the probability of drawing a hitting path $i \leadsto j$ is
\begin{equation}
\text{P}_{\text{h}}(s=i,e=j) = z_{ij}^{\text{h}}/ \displaystyle \sum\limits_{i',j'=1}^{n} z_{i'j'}^{\text{h}}
\label{Eq_hitting_paths_probabilities01}
\end{equation}
with $z_{ij}^{\text{h}} = z_{ij}/z_{jj}$. The partition function for the bag-of-hitting-paths is therefore
\begin{equation}
\mathcal{Z}_{\text{h}} = \displaystyle \sum\limits_{i,j=1}^{n} z_{ij}^{\text{h}} = \displaystyle \sum\limits_{i,j=1}^{n} \frac{z_{ij}}{z_{jj}}
\end{equation}

More information about the bag-of-hitting-paths model can be found in \cite{KEVIN}. Let us simply mention that it can further be shown that the variables $z^{\text{h}}_{ij}$ are defined as
\begin{equation}
 z^{\text{h}}_{ij} = \sum_{\wp\in\mathcal{P}^{h}_{ij}} \tilde{\pi}^{\text{ref}}(\wp) \exp[-\theta \tilde{c}(\wp)] 
\end{equation}
where $\mathcal{P}^{\text{h}}_{ij}$ is now the set of hitting (or absorbing) paths from $i$ to $j$. Finally, it was also shown in \cite{KEVIN} that $z_{ij}^{\text{h}}$ can be interpreted as either:
\begin{itemize}
		\item The expected reward endorsed by an agent (the reward along a path $\wp$ being defined as $\exp[-\theta \tilde{c}(\wp)]$) when traveling from $i$ to $j$ along all possible paths $\wp\in\mathcal{P}^{h}_{ij}$ with probability $\tilde{\pi}^{\text{ref}}(\wp)$.
		\item The expected number of passages through node $j$ for a evaporating random walker starting in node $i$ and walking according to the sub-stochastic transition probabilities $p_{ij}^{\text{ref}} \exp[-\theta c_{ij}]$.
	\end{itemize}

\section{Related work}
\label{RelW}

Graph-based semi-supervised classification has been the subject of intensive research in recent years and a wide range of approaches has been developed in order to tackle the problem \cite{Abney-2008,Zhu-2009,Chapelle-2006,Zhu-2008}: Random-walk-based methods \cite{Zhou-04,Szummer-01}, spectral methods \cite{Chapelle-2002,Kapoor-2005}, regularization frameworks \cite{Zhou-2005,Belkin-2004,Wang-2009,Zhou-2003}, transductive and spectral SVM \cite{joachims03}, to name a few.
We will compare our method (the BoP) to some of those techniques, namely,
	\begin{enumerate}
		\item A simple alignment with the regularized laplacian kernel (RL) based on a sum of similarities, $\mathbf{K}\mathbf{y}_{c}$, where $\mathbf{K} = (\textbf{I}+\lambda \textbf{L})^{-1}$, $\textbf{L} = \textbf{D}- \textbf{A} $ is the laplacian matrix, $\textbf{I}$ is the identity matrix, $\textbf{D}$ is the generalized outdegree matrix, and $\textbf{A}$ is the adjacency matrix of $G$ \cite{Belkin-2004b,Belkin-2004,Kato-2009}. The similarity is computed for each class $c$ in turn. Then, each node is assigned to the class showing the largest similarity. The (scalar) parameter $\lambda > 0$ is the regularization parameter \cite{FoussKernelNN-2012,Mantrach-2011}.
		\item A simple alignment with the regularized normalized laplacian kernel (RNL) based on a sum of similarities,  $\mathbf{K}\mathbf{y}_{c}$, where $\mathbf{K} = (\textbf{I}+\lambda \widetilde{\textbf{L}})^{-1}$, and $\widetilde{\textbf{L}} =$ $ \textbf{D}^{-1/2} \textbf{L} \textbf{D}^{-1/2}$ is the normalized laplacian matrix \cite{Zhou-2003,Zhou-2006}. The assignment to the classes is the same than previous method. The regularized normalized laplacian approach seems less sensitive to the priors of the different classes than the un-normalized regularized laplacian approach (RL) \cite{Zhou-2006}.
		\item A simple alignment with the regularized commute time kernel (RCT) based on a sum of similarities, $\mathbf{K}\mathbf{y}_{c}$, with $\mathbf{K}=(\textbf{D}+\alpha \textbf{A})^{-1}$ \cite{Zhou-2003,FoussKernelNN-2012}. The assignment to the classes is the same as for previous methods. The element $(i,j)$ of this kernel can be interpreted as the discounted cumulated probability of visiting node $j$ when starting from node $i$. The (scalar) parameter $\alpha \in \left] 0,1 \right]$ corresponds to an evaporating or killing random walk where the random walker has a $(1-\alpha$) probability of disappearing at each step. This method provided the best results in a recent comparative study on semi-supervised classification  \cite{FoussKernelNN-2012}.
		\item The harmonic function (HF) approach \cite{Zhu-2003,Abney-2008}, is closely related to the regularization framework of RL and RNL. It is based on a structural contiguity measure that smoothes the predicted values and leads to a model having interesting interpretations in terms of electrical potential and absorbing probabilities in a Markov chain.
		\item The random walk with restart (RWWR) classifier \cite{Pan-2004,Tong-2006,Tong-2008} relies on random walks performed on the weighted graph seen as a Markov chain. More precisely, a group betweenness measure is derived for each class, based on the stationary distribution of a random walk restarting from the labeled nodes belonging to a class of interest. Each unlabeled node is then assigned to the class showing maximal betweenness. In this version \cite{FoussKernelNN-2012}, the random walker has a probability ($1-\alpha$) to be teleported -- with a uniform probability -- to a node belonging to the class of interest $c$.
		\item The discriminative random walks approach ($\mathcal{D}$-walks or DW1; see \cite{Callut-2007}) also relies on random walks performed on the weighted graph seen as a Markov chain. As for the RWWR, a group betweenness measure, based on passage times during random walks, is derived for each class. However, this time, the group betweenness is computed between two groups of nodes and not a single class as for the RWWR method. More precisely, a $\mathcal{D}$-walks is a random walk starting in a labeled node and ending when any node having the same label (possibly the starting node itself) is reached for the first time. During this random walk, the number of visits to any unlabeled node is recorded and corresponds to a group betweenness measure. As for the previous method, each unlabeled node is then assigned to the class showing maximal betweenness.
		\item A modified version of the $\mathcal{D}$-walks (or DW2). The only difference is that all elements of the transition matrix $\mathbf{P}^{\text{ref}}$ (since the random walks is seen as a Markov chain) are multiplied by $\alpha \in \left]0,1 \right]$ so that $\alpha$ can be seen as a probability of continuing the random walk at each time step (and so $(1-\alpha) \in \left[0,1\right[$ is the probability at each step of stopping the random walk. This defines a killing random walk since $\alpha \mathbf{P}^{\text{ref}}$ is now sub-stochastic.
	\end{enumerate}
	
All these methods (see Table \ref{tab:summaryK} for a summary) will be compared to the bag-of-paths (BoP) developed in the next sections.
The random-walk-based methods usually suffer from the fact that the random walker takes too long -- and thus irrelevant -- paths into account so that popular entries are intrinsically favored \cite{Liben-Nowell-2007,Brand-03}. The bag-of-path approach tackles this issue by putting a negative exponential term in  (\ref{W}) and part of its success can be imputed to this fact.

Some authors also considered bounded (or truncated) walks \cite{Mantrach-2011,Sarkar2007,Callut-2008} and obtained promising results on large graphs. This approach could also be considered in our framework in order to tackle large networks; this will be investigated in further work.

Tong et al. suggested a method avoiding to take the inverse of a $n \times n$ matrix for computing the random walk with restart measure \cite{Tong-2006}. They reduce the computing time by partitioning the input graph into smaller communities. Then, a sparse approximate of the random walk with restart is obtained by applying a low rank approximation. This approach suffers from the fact that it adds a hyperparameter $k$ (the number of communities) that depends on the network and is still untractable for large graphs with millions of nodes. On the other hand, the computing time is reduced by this same factor $k$. This is another path to investigate in further work.

Herbster et al. \cite{Herbster-08} proposed a technique for fast label prediction on graphs through the approximation of the graph with either a minimum spanning tree or a shortest path tree. Once the tree has been extracted, the pseudoinverse of the laplacian matrix can be computed efficiently. The fast computation of the pseudo-inverse enables to address prediction problems on large graphs.
Finally, Tang and Liu have investigated relational learning via latent social dimensions \cite{Tang-2009,Tang-2009b,Tang-2010}. They proposed to extract latent social dimensions based on network information (such as Facebook, Twitter,...) first, then they used these as features for discriminative learning (via a SVM for example \cite{Tang-2009}). Their approach tackles very large networks and provides promising results, especially when only a few labeled data are available.\\
We also defined a group betweenness using Freeman's, or shortest path, betweenness \cite{Freeman-1977} and a modified version of Newman's betweenness \cite{Newman-05}. For this last one, the transition probabilities were set to $\mathbf{P}^{\text{ref}}$, and the ending node of the walk was forced to be absorbing. Then, the expected number of visits to each node was recorded and cumulated for each input-output path. However, our BoP group betweenness outperformed these two other class betweenness measure (Consequently, results are not reported in this paper).

\section{The bag-of-paths betweennesses}
\label{BoP}

In order to define the BoP classifier, we need to introduce the  BoP group betweenness centrality. This concept is itself an extension of the BoP betweenness centrality, which will be developed in the next subsection.
This section starts with the BoP betweenness centrality concept in Subsection \ref{aaa}. Then, its extension, the BoP group betweenness centrality, is described in Subsection \ref{bbb}.


\subsection{The bag-of-paths betweenness centrality}
\label{aaa}

The BoP betweenness measure quantifies to which extent a node $j$ lies in between other pairs of nodes $i,k$, and therefore is an important intermediary between nodes.
Recall that from (\ref{Eq_bag_of_paths_probabilities01}) the probability of drawing a path starting at node $i$ ($s=i$) and ending in node $k$ ($e=k$) from a regular bag-of-paths is $
\text{P}(s=i,e=k) = z_{ik}/\mathcal{Z}$.

We now compute the probability $\text{P}(s=i,int=j,e=k;i \ne j \ne k \ne i)$ -- or $\text{P}_{ijk}$ in short -- that such paths visit an \emph{intermediate} node $int = j$ when $i \ne j \ne k \ne i$. Indeed, from (\ref{Boltzmann}),
\begin{equation}
\text{P}_{ijk}
= \frac{{\displaystyle \sum_{\wp\in\mathcal{P}_{ik}}} \delta(j \in \wp) \, \tilde{\pi}^{\text{ref}}(\wp)\exp\left[-\theta  \tilde{c}(\wp)\right]}{{\displaystyle \sum_{\wp' \in \mathcal{P}}} \tilde{\pi}^{\text{ref}}(\wp')\exp\left[-\theta  \tilde{c}(\wp')\right]}
\label{Eq_bag_of_paths_probability_intermediate01}
\end{equation}where $\delta(j \in \wp)$ is the indicator function, i.e. is equal to 1 if the path $\wp$ contains (at least once) node $j$, and 0 otherwise.

We will now use the fact that each path $\wp_{ik}$ between $i$ and $k$ passing through $j$ can be decomposed uniquely into a \emph{hitting} sub-path $\wp_{ij}$ from $i$ to $j$ and a regular sub-path $\wp_{jk}$ from $j$ to $k$. The sub-path $\wp_{ij}$ is found by following path $\wp_{ik}$ until reaching $j$ for the first time\footnote{This is the reason why we introduced \emph{hitting paths}.}. Therefore, for $i \ne j \ne k \ne i$,
\begin{align}
\text{P}_{ijk}
=& \frac{1}{\mathcal{Z}} {\displaystyle \sum_{\wp\in\mathcal{P}_{ik}}} \delta(j \in \wp) \, \tilde{\pi}^{\text{ref}}(\wp)\exp\left[-\theta  \tilde{c}(\wp)\right] \nonumber \\
=& \frac{1}{\mathcal{Z}} {\displaystyle \sum_{\wp_{ij} \in \mathcal{P}^{\text{h}}_{ij}} \sum_{\wp_{jk} \in \mathcal{P}_{jk}} } \tilde{\pi}^{\text{ref}}(\wp_{ij}) \tilde{\pi}^{\text{ref}}(\wp_{jk}) \nonumber  \\ & \times \exp\left[-\theta  \tilde{c}(\wp_{ij})  \right] \exp\left[-\theta  \tilde{c}(\wp_{jk}) \right] \nonumber \\
=& \frac{1}{\mathcal{Z}} \left[ {\displaystyle \sum_{\wp_{ij} \in \mathcal{P}^{\text{h}}_{ij}} } \tilde{\pi}^{\text{ref}}(\wp_{ij}) \exp\left[-\theta  \tilde{c}(\wp_{ij})  \right] \right] \nonumber \\
& \times \left[ {\displaystyle \sum_{\wp_{jk} \in \mathcal{P}_{jk}} }  \tilde{\pi}^{\text{ref}}(\wp_{jk}) \exp\left[-\theta  \tilde{c}(\wp_{jk}) \right] \right]  \nonumber \\
=& \mathcal{Z}_{\text{h}} \frac{ \left[ {\displaystyle \sum_{\wp_{ij} \in \mathcal{P}^{\text{h}}_{ij}} } \tilde{\pi}^{\text{ref}}(\wp_{ij}) \exp\left[-\theta  \tilde{c}(\wp_{ij})  \right] \right] } {\mathcal{Z}_{\text{h}}}  \nonumber \\
& \times \frac{\left[ {\displaystyle \sum_{\wp_{jk} \in \mathcal{P}_{jk}} }  \tilde{\pi}^{\text{ref}}(\wp_{jk}) \exp\left[-\theta  \tilde{c}(\wp_{jk}) \right] \right] }{\mathcal{Z}} \nonumber \\
=& \mathcal{Z}_{\text{h}} \, \text{P}_{\text{h}}(s=i,e=j) \, \text{P}(s=j,e=k), \text{ for } i \ne j \ne k \ne i
\label{Eq_bag_of_paths_probability_intermediate02}
\end{align}

Thus, from s (\ref{BagOfP}), (\ref{Eq_hitting_paths_probabilities01}),
\begin{align}
\text{P}_{ijk} &= \mathcal{Z}_{\text{h}} \, \text{P}_{\text{h}}(s=i,e=j) \, \text{P}(s=j,e=k) \nonumber \\
&= \frac{\mathcal{Z}_{\text{h}} \left( \dfrac{z_{ij}} {z_{jj}} \right) \left( z_{jk} \right)}{\mathcal{Z}_{\text{h}} \mathcal{Z}} \nonumber \\
&= \frac{1}{\mathcal{Z}}  \frac{z_{ij} z_{jk}}{z_{jj}}, \text{ for } i \ne j \ne k \ne i.
\end{align}

Since $\text{P}(s=i,int=j,e=k)$ is only meaningful when $i \ne j \ne k \ne i$, from  (\ref{Eq_bag_of_paths_probability_intermediate01}) and (\ref{Eq_bag_of_paths_probability_intermediate02}), since we are only interested in the case in which this condition is false:
\begin{align}
&\text{P}(s=i,int=j,e=k; i \ne j \ne k) \nonumber \\
&= \frac{1}{\mathcal{Z}}  \frac{z_{ij} z_{jk}}{z_{jj}} \, \delta(i \ne j \ne k)
\label{Eq_bag_hitting_paths_a_posteriori17}
\end{align}

Now, the \emph{a posteriori} probabilities of visiting intermediate node $j$ given that the path starts in $i$ and ends in $k$ are therefore (remember that $i \ne j \ne k \ne i$)
\begin{align}
&\text{P}(int=j|s=i,e=k; i \ne j \ne k \ne i) \nonumber \\
&= \frac{\text{P}(s=i,int=j,e=k; i \ne j \ne k \ne i)}{\displaystyle \sum_{j'=1}^{n} \text{P}(s=i, int=j',e=k; i \ne j' \ne k \ne i)} \nonumber \\
&= \frac{\left( \dfrac{z_{ij} z_{jk}}{z_{jj} \mathcal{Z}} \right)} {\displaystyle \left( \sum_{j'=1}^{n} \dfrac{z_{ij'} z_{j'k}}{z_{j'j'} \mathcal{Z}} \, \delta(i \ne j' \ne k \ne i)  \right)} \delta(i \ne j \ne k \ne i) \nonumber \\
&= \frac{\left( \dfrac{z_{ij} z_{jk}}{z_{jj}} \right)} {\displaystyle \sum_{\substack{j'=1\\j' \notin \{ i,k \}}}^{n} \left( \dfrac{z_{ij'} z_{j'k}}{z_{j'j'}} \right) } \delta(i \ne j \ne k \ne i)
\label{Eq_bag_hitting_paths_a_posteriori01}
\end{align}
where we assumed that the node $k$ can be reached from node $i$ and we used  (\ref{Eq_bag_hitting_paths_a_posteriori17}).

Based on these a posteriori probabilities, the bag-of-paths betweenness of node $j$ is defined as the sum of the a posteriori probabilities of visiting $j$ for all possible starting-destination pairs $i,k$:
\begin{equation}
\begin{aligned}
\text{bet}_{j} 
&= \sum_{i=1}^{n} \sum_{k=1}^{n} \text{P}(int=j|s=i,e=k; i \ne j \ne k \ne i) \\
&= \frac{1}{z_{jj}} \sum_{\substack{i=1\\i \ne j}}^{n} \sum_{\substack{k=1\\k \notin \{i,j\}}}^{n} \dfrac{z_{ij} z_{jk}}{ \displaystyle \sum_{\substack{j'=1\\j' \notin \{i,k\}}}^{n} \left( \dfrac{z_{ij'} z_{j'k}}{z_{j'j'}} \right) }
\end{aligned}
\label{Eq_a_posteriori_bag_hitting_paths01}
\end{equation}

This quantity indicates to which extent a node $j$ lies in between pairs of nodes, and therefore to which extent $j$ is an important intermediary in the network.

Let us now derive the matrix formula computing the betweenness vector $\mathbf{bet}$. This vector contains the $n$ betweennesses for each node. First of all, the normalization factor will be computed, $n_{ik} = \sum_{j'=1}^{n}~(1-\delta_{ij'}) (1-\delta_{j'k})~(z_{ij'} z_{j'k})/z_{j'j'}$, appearing in the denominator of  (\ref{Eq_a_posteriori_bag_hitting_paths01}). We easily see that $n_{ik}~=~\sum_{j'=1}^{n}~\{(1-\delta_{ij'})z_{ij'}\} \{1/z_{j'j'}\}$$\{(1-\delta_{j'k})z_{j'k}\}$.~ Therefore, by defining $\mathbf{D}_{\text{z}}^{-1} = (\mathbf{Diag}(\mathbf{Z}))^{-1}$ whose diagonal contains elements $1/z_{j'j'}$, the matrix containing the normalization factors $n_{ik}$ is $\mathbf{N}=(\mathbf{Z} - \mathbf{Diag}(\mathbf{Z})) \mathbf{D}_{\text{z}}^{-1}(\mathbf{Z} - \mathbf{Diag}(\mathbf{Z}))$.~

Moreover, the term  $\sum_{i=1}^{n} \sum_{k=1}^{n} \delta(i \ne j \ne k \ne i) z_{ij} (1/n_{ik}) z_{jk}$ appearing in  (\ref{Eq_a_posteriori_bag_hitting_paths01}) can be rewritten as $\sum_{i=1}^{n} \sum_{k=1}^{n} \{(1-\delta_{ji}) z_{ji}^{\text{t}}\} \{(1-\delta_{ik}) (1/n_{ik})\} \{(1-\delta_{kj}) z_{kj}^{\text{t}}\}$ where $z_{ij}^{\text{t}}$ is the element $i$, $j$ of matrix $\mathbf{Z}^{\text{T}}$ (the transpose of $\mathbf{Z}$). In matrix form, $\mat{bet}$ is therefore equal to
\begin{align}
\mathbf{bet} = 
&\mathbf{D}_{\text{z}}^{-1} \mathbf{diag} [ (\mathbf{Z}^{\text{T}} - \mathbf{Diag}(\mathbf{Z})) \nonumber \\
&\cdot (\mathbf{N}^{\div} - \mathbf{Diag}(\mathbf{N}^{\div})) (\mathbf{Z}^{\text{T}} - \mathbf{Diag}(\mathbf{Z})) ]
\end{align}
where matrix $\mathbf{N}^{\div}$ contains elements $n_{ik}^{\div} = 1/n_{ik}$ with $\mathbf{N} = (\mathbf{Z} - \mathbf{Diag}(\mathbf{Z})) \mathbf{D}_{\text{z}}^{-1} (\mathbf{Z} - \mathbf{Diag}(\mathbf{Z}))$. Moreover, for a given matrix $\mathbf{M}$, $\mathbf{diag}(\mathbf{M})$ is a column vector containing the diagonal of $\mathbf{M}$ while $\mathbf{Diag}(\mathbf{M})$ is a diagonal matrix containing the diagonal of $\mathbf{M}$.

	
\subsection{The bag-of-paths group betweenness centrality}
\label{bbb}

Let us now generalize the bag-of-paths betweenness to a group betweenness measure. Quite naturally, the \textbf{bag-of-paths group betweenness} of node $j$ will be defined as
\begin{equation}
\text{gbet}_{j}(\mathcal{C}_i,\mathcal{C}_k) = \text{P}(int=j|s \in \mathcal{C}_i,e \in \mathcal{C}_k ; s \ne int \ne e \ne s)
\label{Eq_a_posteriori_bag_hitting_paths_group01}
\end{equation}and can be interpreted as the extent to which the node $j$ lies in between the two sets of nodes $\mathcal{C}_i$ and $\mathcal{C}_k$. It is assumed that the sets $\mathcal{C}_i$,$(i=1...m$) are disjoint. Using Bayes' law provides
\begin{align}
&\text{P}(int=j|s \in \mathcal{C}_i,e \in \mathcal{C}_k; s \ne int \ne e \ne s) \nonumber \\
 &= \frac{\text{P}(s \in \mathcal{C}_i,int=j,e \in \mathcal{C}_k; s \ne int \ne e \ne s)}{\text{P}(s \in \mathcal{C}_i,e \in \mathcal{C}_k; s \ne int \ne e \ne s)} \nonumber \\
&= \dfrac{\displaystyle \sum_{i' \in \mathcal{C}_i} \sum_{k' \in \mathcal{C}_k} \text{P}(s=i',int=j,e=k'; s \ne int \ne e \ne s)}{\displaystyle \sum_{j'=1}^{n} \sum_{i' \in \mathcal{C}_i} \sum_{k' \in \mathcal{C}_k} \text{P}(s=i',int=j',e=k'; s \ne int \ne e \ne s)}
\label{Eq_a_posteriori_bag_hitting_paths_bayes01}
\end{align}Substituting  (\ref{Eq_bag_hitting_paths_a_posteriori01}) for the probabilities in  (\ref{Eq_a_posteriori_bag_hitting_paths_bayes01}) allows to compute the group betweenness measure in terms of the elements of the fundamental matrix $\mathbf{Z}$:
\begin{align}
\text{gbet}_{j}(\mathcal{C}_i,\mathcal{C}_k)
&=  \dfrac{\dfrac{1}{\mathcal{Z}} \displaystyle \sum_{i' \in \mathcal{C}_i} \sum_{k' \in \mathcal{C}_k} \delta(i' \ne j \ne k' \ne s) \frac{z_{i'j} z_{jk'}}{z_{jj}}}{\displaystyle \dfrac{1}{\mathcal{Z}} \sum_{j'=1}^{n} \sum_{i' \in \mathcal{C}_i} \sum_{k' \in \mathcal{C}_k} \delta(i' \ne j' \ne k' \ne s) \frac{z_{i'j'} z_{j'k'}}{z_{j'j'}} } \nonumber \\
&=  \dfrac{ \displaystyle \frac{1}{z_{jj}} \sum_{i' \in \mathcal{C}_i} \sum_{k' \in \mathcal{C}_k} \delta(i' \ne j \ne k' \ne s) \, z_{i'j} z_{jk'}}{\displaystyle  \sum_{j'=1}^{n} \sum_{i' \in \mathcal{C}_i} \sum_{k' \in \mathcal{C}_k} \delta(i' \ne j' \ne k' \ne s) \frac{z_{i'j'} z_{j'k'}}{z_{j'j'}} }
\label{Eq_bag_of_paths_group_betweenness01}
\end{align}
where the denominator is simply a normalization factor ensuring that the probability distribution sums to one. It is therefore sufficient to compute the numerator and then normalize the resulting quantity.

Let us put this expression in matrix form. We first define $\mathbf{D}_{\text{z}} = \mathbf{Diag}(\mathbf{Z})$ ($\mathbf{D}_{\text{z}}$ is just the matrix $\mathbf{Z}$ where all non-diagonal elements are set to zero) and $z_{ij}^{\text{t}}$ as element $i,j$ of matrix $\mathbf{Z}^{\text{T}}$ (the transpose of $\mathbf{Z}$). Here again, it is assumed that node $i'$ and $k'$ belong to different sets, $\mathcal{C}_i \ne \mathcal{C}_k$, so that $i'$ and $k'$ are necessarily different. Therefore, if $\mathbf{y}_k$ is a binary membership vector indicating which node belongs to class $\mathcal{C}_k$ (as described in Section \ref{Notations}), the numerator of  (\ref{Eq_bag_of_paths_group_betweenness01}) can be rewritten as (remembering that $\mathcal{C}_i\neq \mathcal{C}_k$)
\begin{align}
&\text{numerator}\left(\text{gbet}_{j}(\mathcal{C}_i,\mathcal{C}_k) \right) \nonumber \\
&= \frac{1}{z_{jj}} \sum_{i' \in \mathcal{C}_i} \sum_{k' \in \mathcal{C}_k} (1 - \delta_{ji'}) (1 - \delta_{jk'}) \, z_{i'j} z_{jk'} \nonumber \\
&= \frac{1}{z_{jj}} \left( \sum_{i' \in \mathcal{C}_i} (1 - \delta_{ji'}) z_{ji'}^{\text{t}} \right) \left( \sum_{k' \in \mathcal{C}_k} (1 - \delta_{jk'}) z_{jk'} \right) \nonumber \\
&= \frac{1}{z_{jj}} \left( \sum_{i'=1}^{n} (1 - \delta_{ji'}) z_{ji'}^{\text{t}} y_{ii'} \right) \left( \sum_{k'=1}^{n} (1 - \delta_{jk'}) z_{jk'} y_{kk'} \right)  \label{nolabel}
\end{align}
Consequently, in matrix form, the group betweenness vector reads
\begin{equation}
\begin{cases}
\begin{aligned}
 \mathbf{gbet}(\mathcal{C}_i,\mathcal{C}_k)
\leftarrow &\; \mathbf{D}_{\text{z}}^{-1} \left( (\mat{Z}_{0}^{\text{T}} \mathbf{y}_{i}) \circ (\mat{Z}_{0} \mathbf{y}_{k}) \right) \\
& \text{ with } \mat{Z}_{0} = \mathbf{Z} - \mathbf{Diag}(\mathbf{Z}),\\
\end{aligned} \\
 \mathbf{gbet}(\mathcal{C}_i,\mathcal{C}_k) \leftarrow \dfrac{\mathbf{gbet}(\mathcal{C}_i,\mathcal{C}_k)}{\|\mathbf{gbet}(\mathcal{C}_i,\mathcal{C}_k)\|_1} \text{ (normalization)} 
\end{cases}
\label{Eq_bag_hitting_paths_group_betweenness_matrix01}
\end{equation}
where $\circ$ is the elementwise multiplication (Hadamard product) and we assume $i \ne k$. In this equation, the vector $\mathbf{gbet}(\mathcal{C}_i,\mathcal{C}_k)$ must be normalized by dividing it by its $L_{1}$ norm. Notice that $\mat{Z}_{0} = \mathbf{Z} - \mathbf{Diag}(\mathbf{Z})$ is simply the fundamental matrix whose diagonal is set to zero.

\section{Semi-supervised classification through the bag-of-paths group betweenness}
\label{BoPClass}

In this section, the bag of hitting paths model will be used for \textbf{classification purposes}. From  (\ref{Eq_a_posteriori_bag_hitting_paths_group01}) and (\ref{Eq_bag_of_paths_group_betweenness01}), recall that the bag-of-paths group betweenness measure was defined as
\begin{align}
\text{gbet}_{j}(\mathcal{C}_i,\mathcal{C}_k)
&= \text{P}(int=j|s \in \mathcal{C}_i,e \in \mathcal{C}_k; s \ne int \ne e) \nonumber \\
&=  \dfrac{ \displaystyle \frac{1}{z_{jj}} \sum_{i' \in \mathcal{C}_i} \sum_{k' \in \mathcal{C}_k} \delta(i' \ne j \ne k') \, z_{i'j} z_{jk'}}{\displaystyle  \sum_{j'=1}^{n} \sum_{i' \in \mathcal{C}_i} \sum_{k' \in \mathcal{C}_k} \delta(i' \ne j' \ne k') \frac{z_{i'j'} z_{j'k'}}{z_{j'j'}} } 
\label{Eq_group betweenness_classification01}
\end{align}
and, as before, the denominator of  (\ref{Eq_group betweenness_classification01}) aims to normalize the probability distribution so that it sums to one. We will therefore compute the numerator of  (\ref{Eq_group betweenness_classification01}) and then normalize the resulting quantity.

Notice, however, that in the derivation of the matrix form of the group betweenness (see  (\ref{Eq_bag_hitting_paths_group_betweenness_matrix01})), it was assumed that $\mathcal{C}_i \ne \mathcal{C}_k$. We will now recompute this quantity when starting and ending in the same class $c$, i.e. calculating $\text{gbet}_{j}(\mathcal{C}_c,\mathcal{C}_c)$. This will provide a measure of the extent to which nodes of $G$ are in between -- and therefore in the neighbourhood of -- class $c$.  A within-class betweenness is thus defined for each class $c$ and each node will be assigned to the class showing the highest betweenness. The main hypothesis underlying this classification technique is that a node is likely to belong to the same class as its ``neighboring nodes". This is usually called the local consistency assumption (also called smoothness and cluster assumption \cite{Zhu-2003,Chapelle-2006,Kolaczyk-2009b}).

The same reasoning as for deriving  (\ref{Eq_bag_hitting_paths_group_betweenness_matrix01}) is applied in order to compute the numerator of  (\ref{Eq_group betweenness_classification01}),
\begin{align}
&\text{numerator}\left(\text{gbet}_{j}(\mathcal{C}_c,\mathcal{C}_c)\right) \nonumber\\
&= \frac{1}{z_{jj}} \sum_{i' \in \mathcal{C}_c} \sum_{k' \in \mathcal{C}_c} \delta(i' \ne j \ne k' \ne i') \, z_{i'j} z_{jk'} \nonumber \\
&= \frac{1}{z_{jj}} \sum_{i' \in \mathcal{C}_c} \sum_{k' \in \mathcal{C}_c} (1 - \delta_{ji'}) (1 - \delta_{i'k'}) (1 - \delta_{jk'}) \, z_{i'j} z_{jk'} \nonumber \\
&= \frac{1}{z_{jj}} \sum_{i' \in \mathcal{C}_c} \sum_{k' \in \mathcal{C}_c} (1 - \delta_{ji'}) (1 - \delta_{jk'}) \, z_{i'j} z_{jk'} \nonumber \\
&\quad - \frac{1}{z_{jj}} \sum_{i' \in \mathcal{C}_c} \sum_{k' \in \mathcal{C}_c} (1 - \delta_{ji'}) \delta_{i'k'} (1 - \delta_{jk'}) \, z_{i'j} z_{jk'} \nonumber \\
&= \frac{1}{z_{jj}} \left( \sum_{i' \in \mathcal{C}_c} (1 - \delta_{ji'}) z_{i'j} \right) \left( \sum_{k' \in \mathcal{C}_c} (1 - \delta_{jk'}) z_{jk'} \right) \nonumber \\
&\quad - \frac{1}{z_{jj}} \sum_{i' \in \mathcal{C}_c} (1 - \delta_{ji'}) z_{i'j} (1 - \delta_{ji'})z_{ji'} \nonumber \\
&= \frac{1}{z_{jj}} \left( \sum_{i'=1}^{n} (1 - \delta_{ji'}) z_{ji'}^{\text{t}} y^c_{i'} \right) \left( \sum_{k'=1}^{n} (1 - \delta_{jk'}) z_{jk'} y^c_{k'} \right) \nonumber \\
&\quad - \frac{1}{z_{jj}} \sum_{i'=1}^{n} ((1 - \delta_{ji'}) z_{ji'}^{\text{t}}) ((1 - \delta_{ji'}) z_{ji'}) \, y^c_{i'}
\end{align}where $y^c_i$ is a binary indicator indicating if node $i$ belongs to the class $c$ (so that $y^c_i$ is equal to $y_{ic}$, the element $(i,c)$ of matrix $\mathbf{Y}$) and $z_{ij}^{\text{t}}$ is the element $i,j$ of matrix $\mathbf{Z}^{\text{T}}$ (the transpose of $\mathbf{Z}$).

Now, we easily observe that $(1 - \delta_{ji'}) z_{ji'}$ is element $(j,i')$ of matrix $\mat{Z}_{0} = \mat{Z} - \mathbf{Diag}(\mat{Z})$. This expression can thus be re-expressed in matrix form as
\begin{align}
\text{numerator}&(\mathbf{gbet}(\mathcal{C}_c,\mathcal{C}_c)) \nonumber \\
&= \mathbf{D}_{\text{z}}^{-1} \big[ (\mat{Z}_{0}^{\text{T}} \mathbf{y}^{c}) \circ (\mat{Z}_{0} \mathbf{y}^{c}) - (\mathbf{Z}_{0}^{\text{T}}  \circ \mathbf{Z}^{0}) \mathbf{y}^{c} \big], \nonumber \\
&\quad \text{ with } \mat{Z}_{0} = \mathbf{Z} - \mathbf{Diag}(\mathbf{Z})
\end{align}
where $\circ$ is the elementwise multiplication (Hadamard product). After having computed this equation, the numerator must be normalized in order to obtain $\mathbf{gbet}(\mathcal{C}_c,\mathcal{C}_c)$ (see  (\ref{Eq_group betweenness_classification01})).

Finally, if we want to classify a node, $\mathbf{gbet}(\mathcal{C}_c,\mathcal{C}_c)$ is computed for each class $c$ in turn and then, for each node, the class showing the maximal betweenness is chosen,
\begin{equation}
\begin{aligned}
&\boldsymbol{\widehat{\ell}}
= \operatorname*{arg\,max}_{c \in \mathcal{L}}
\left( \mathbf{gbet}(\mathcal{C}_{c},\mathcal{C}_{c}) \right),\text{ with } \\
&\begin{cases}
\mat{Z}_{0} \leftarrow \mathbf{Z} - \mathbf{Diag}(\mathbf{Z}) \text{ (set diagonal to 0)} \\
\mathbf{gbet}(\mathcal{C}_{c},\mathcal{C}_{c}) \leftarrow \mathbf{D}_{\text{z}}^{-1} \big[ (\mat{Z}_{0}^{\text{T}} \mathbf{y}^{c}) \circ (\mat{Z}_{0} \mathbf{y}^{c}) - (\mathbf{Z}_{0}^{\text{T}}  \circ \mathbf{Z}_{0}) \mathbf{y}^{c} \big] \\
\mathbf{gbet}(\mathcal{C}_c,\mathcal{C}_c) \leftarrow \dfrac{\mathbf{gbet}(\mathcal{C}_c,\mathcal{C}_c)}{\|\mathbf{gbet}(\mathcal{C}_c,\mathcal{C}_c)\|_1} \text{ (normalization) }
\end{cases}
\end{aligned}
\end{equation}

The pseudo-code for the BoP classifier can be found in Algorithm 1. Of course, once computed, the group betweenness is only used for the unlabeled nodes.

\begin{algorithm}[th!]


\begin{algorithmic}[1]
\caption{Classification through the bag-of-paths group betweenness algorithm.}
\small
\REQUIRE $\,$ \\
 -- A weighted directed graph $G$ containing $n$ nodes, represented by its $n\times n$ adjacency matrix $\mathbf{A}$, containing affinities. \\
 -- The $n\times n$ cost matrix $\mathbf{C}$ associated to $G$ (usually, the costs are the inverse of the affinities, but other choices are possible).\\
 -- $m$ binary indicator vectors $\mathbf{y}_{c}$ containing as entries $1$ for nodes belonging to the class whose label index is $c$, and $0$ otherwise. Classes are mutually exclusive. \\
 -- The inverse temperature parameter $\theta$.\\
 
\ENSURE $\,$ \\
 -- The $n \times m$ membership matrix $\mathbf{U}$ containing the membership
of each node $i$ to class $k$, $u_{ik}$.\\

\STATE $\mathbf{D} \leftarrow \mathbf{Diag}(\mathbf{A}\mathbf{e})$ \COMMENT{the row-normalization matrix} \\
\STATE $\mathbf{P}^{\text{ref}} \leftarrow \mathbf{D}^{-1} \mathbf{A}$ \COMMENT{the reference transition probabilities matrix} \\
\STATE $\mathbf{W} \leftarrow \mathbf{P}^{\text{ref}}\circ\exp\left[-\theta  \mathbf{C}\right]$ \COMMENT{elementwise exponential and multiplication $\circ$} \\
\STATE $\mathbf{Z} \leftarrow (\mathbf{I}-\mathbf{W}\mathbf{)}^{-1}$ \COMMENT{the fundamental matrix} \\
\STATE $\mat{Z}_{0} \leftarrow \mathbf{Z} - \mathbf{Diag}(\mathbf{Z})$ \COMMENT{set diagonal to zero} \\
\STATE $\mathbf{D}_{\text{z}} \leftarrow \mathbf{Diag}(\mathbf{Z})$ \\
\STATE $\mathbf{U} \leftarrow \mathbf{Zeros}(n,m)$ \COMMENT{initialize the membership matrix}
\FOR{$c=1$ to $m$}
\STATE $\widehat{\mathbf{y}}^{*}_{c}
\leftarrow \mathbf{D}_{\text{z}}^{-1} \big[ (\mat{Z}_{0}^{\text{T}} \mathbf{y}^{c}) \circ (\mat{Z}^{0} \mathbf{y}^{c}) - (\mathbf{Z}_{0}^{\text{T}}  \circ \mathbf{Z}_{0}) \mathbf{y}^{c} \big]$ \COMMENT{compute the group betweenness for class $c$; $\circ$ is the elementwise multiplication (Hadamard product)}
\STATE $\widehat{\mathbf{y}}^{*}_{c} \leftarrow \dfrac{\widehat{\mathbf{y}}^{*}_{c}}{\|\widehat{\mathbf{y}}^{*}_{c}\|_1}$ \COMMENT{normalize the betweenness scores}
\ENDFOR
\STATE $\boldsymbol{\widehat{\ell}} \leftarrow {\displaystyle \operatorname*{arg\,max}_{c \in \mathcal{L}}} (\widehat{\mathbf{y}}^{*}_{c})$ \COMMENT{each node is assigned to the class showing the largest class betweenness}
\FOR{$i=1$ to $n$}
\STATE $u_{i,\widehat{\ell}_{i}} \leftarrow 1$ \COMMENT{compute the elements of the membership matrix}
\ENDFOR
\RETURN $\mathbf{U}$

\end{algorithmic} 

\label{Alg_bag_of_hitting_paths_potential_classification01} 
\end{algorithm}

\section{Experimental comparisons}
\label{Exp}

In this section, the bag-of-paths group betweenness approach for semi-supervised classification (referred to as the BoP classifier for simplicity) will be compared to other semi-supervised classification techniques on multiple data sets. The different classifiers to which the BoP classifier will be compared were already introduced in Section \ref{RelW} and are recalled in Table \ref{tab:summaryK}. 

The goal of the experiments of this section is to classify unlabeled nodes in medium-size partially labeled graphs and to compare the different methods in terms of classification accuracy. This comparison is performed on medium-size networks only since kernel approaches are difficult to compute on large networks. The computational tractability of the methods used in this experimental section will also be analyzed.

This section is organized as follows. First, the data sets used for the semi-supervised classification will be described in Subsection \ref{DATA}. Second, the experimental methodology is detailed in Subsection \ref{ExpM}. Third, the results will be discussed in Subsection \ref{ResDis}. Finally, the computation time will be investigated in Subsection \ref{CTime}.

\begin{table}[htdp]
\caption{Class distribution of the \emph{IMDb-proco} data set.}
\scriptsize
\begin{center}
\begin{tabular}{lccc}\hline
\textbf{Class} & \textbf{IMDb}\\
\hline
\\
High-revenue&572\\
Low-revenue&597\\
\\
\textbf{Total} &1169\\
\hline
\end{tabular}
\label{tab: TabIMDb}
\end{center}
\vspace{-0.5cm}
\end{table}

\begin{table*}[htdp]
\centering
\caption{Class distribution of the nine \emph{Newsgroups} data sets. NewsGroup 1-3 contain two classes, NewsGroup 4-6 contain three classes and NewsGroup 7-9 contain five classes.}
\scriptsize
\begin{center}
\begin{tabular}{cccccccccc}\hline
{\textbf{Class}} & {\textbf{NG1}}& {\textbf{NG2}}& {\textbf{NG3}}& {\textbf{NG4}}& {\textbf{NG5}}& {\textbf{NG6}}& {\textbf{NG7}}& {\textbf{NG8}}& {\textbf{NG9}}\\
\hline
\\
1&200&198&200&200&200&197&200&200&200\\
2&200&200&199&200&198&200&200&200&200\\
3&   &   &   &200&200&198&200&198&197\\
4&   &   &   &   &   &   &200&200&200\\
5&   &   &   &   &   &   &198&200&200\\
\\
{\textbf{Total}} &400&398&399&600&598&595&998&998&997\\
\hline
\end{tabular}
\label{tab: TabNG}
\end{center}
\end{table*}


\begin{table}[htdp]
\caption{Class distribution of the four \emph{WebKB cocite} data sets.}
\scriptsize
\begin{center}
\begin{tabular}{lcccc}\hline
\textbf{Class} & \textbf{Cornell}& \textbf{Texas}& \textbf{Washington}& \textbf{Wisconsin}\\
\hline
\\
Course&54&51&170&83\\
Department&25&36&20&37\\
Faculty&62&50&44&37\\
Project&54&28&39&25\\
Staff&6&6&10&11\\
Student&145&163&151&155\\
\\
\textbf{Total} &346&334&434&348\\
\textbf{Majority} &&&&\\
\textbf{class (\%)} &41.9&48.8&39.2&44.5\\
\hline
\end{tabular}
\label{tab: Tabcocite}
\end{center}
\end{table}

%
%
\begin{figure*}[t!]
	\centering
		\includegraphics[width=19cm]{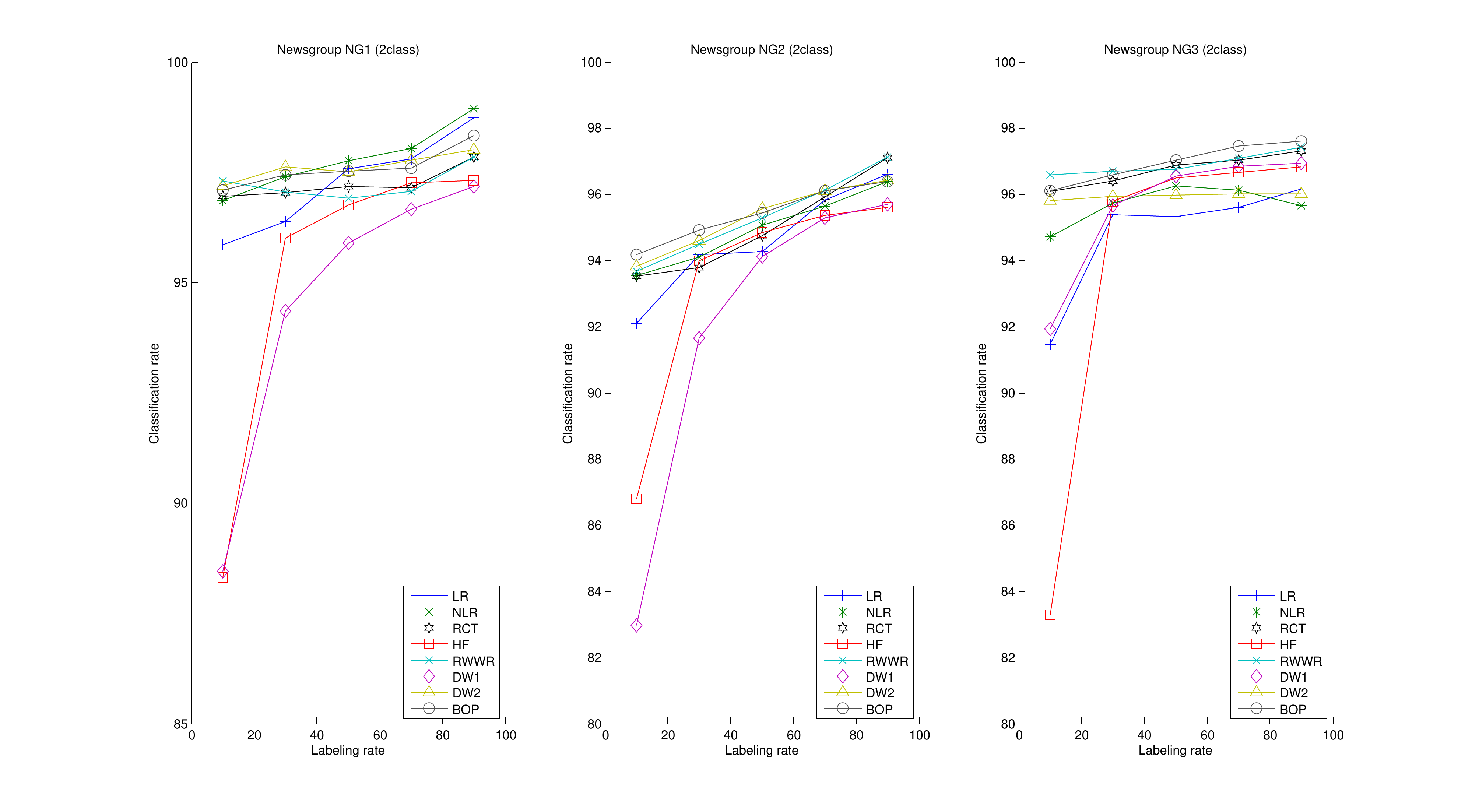}
		\caption{Classification rates in percents, averaged over 20 runs, obtained on partially labeled graphs. Results are reported for the eight methods (RL, RNL, RCT, HF, RWWR, DW1, DW2, BoP) and for five labeling rates (10\%, 30\%, 50\%, 70\%, 90\%). These graphs show the results obtained on the three 2-classes \emph{Newsgroups} data sets.}
		\label{fig: NG2}
		\end{figure*}
		
			\begin{figure*}[t!]
			\centering
		\includegraphics[width=19cm]{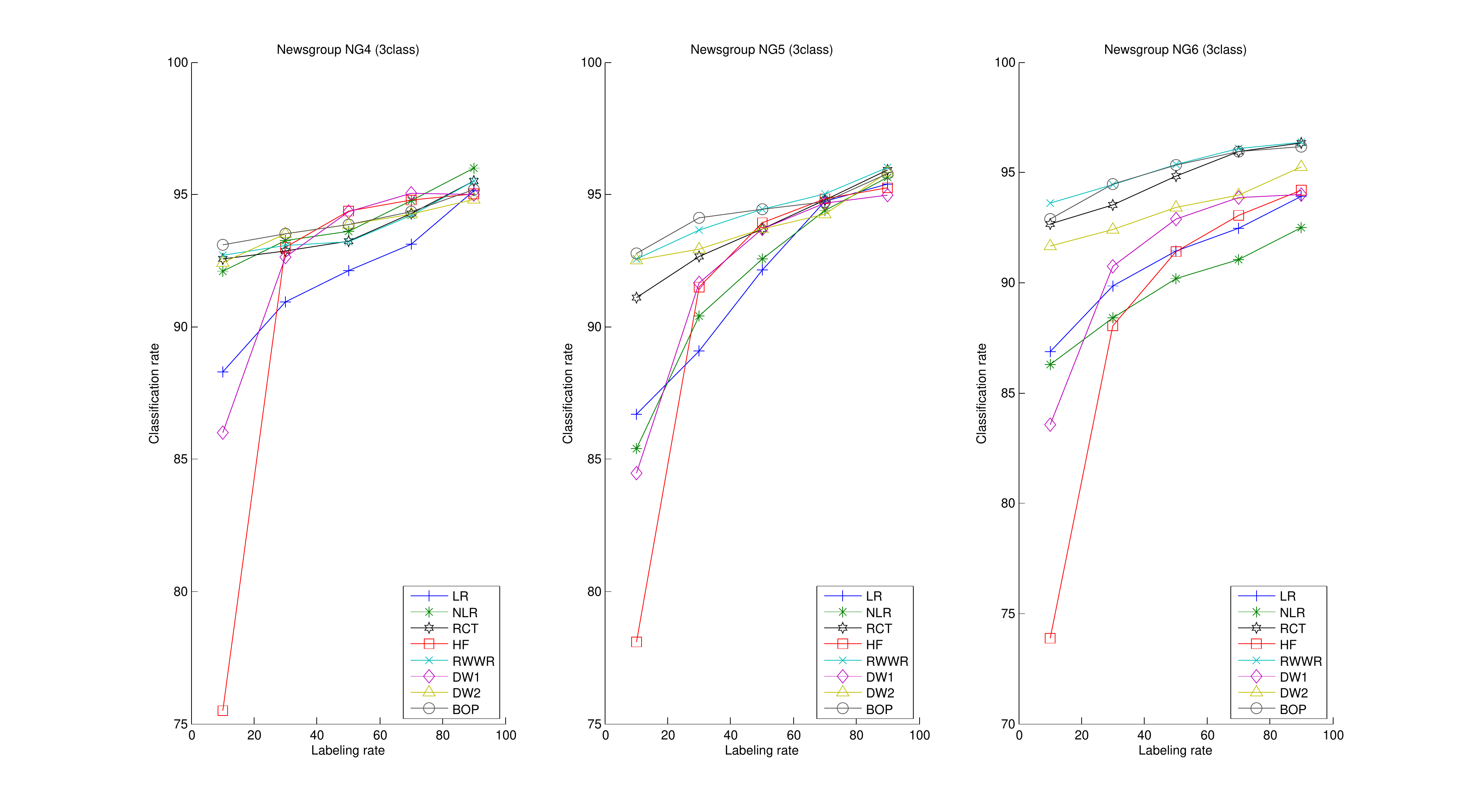}
		\caption{Classification rates in percents, averaged over 20 runs, obtained on partially labeled graphs. Results are reported for the eight methods (RL, RNL, RCT, HF, RWWR, DW1, DW2, BoP) and for five labeling rates (10\%, 30\%, 50\%, 70\%, 90\%). These graphs show the results obtained on the three 3-classes \emph{Newsgroups} data sets.}
		\label{fig: NG3}
		\end{figure*}
		
			\begin{figure*}[t!]
			\centering
		\includegraphics[width=19cm]{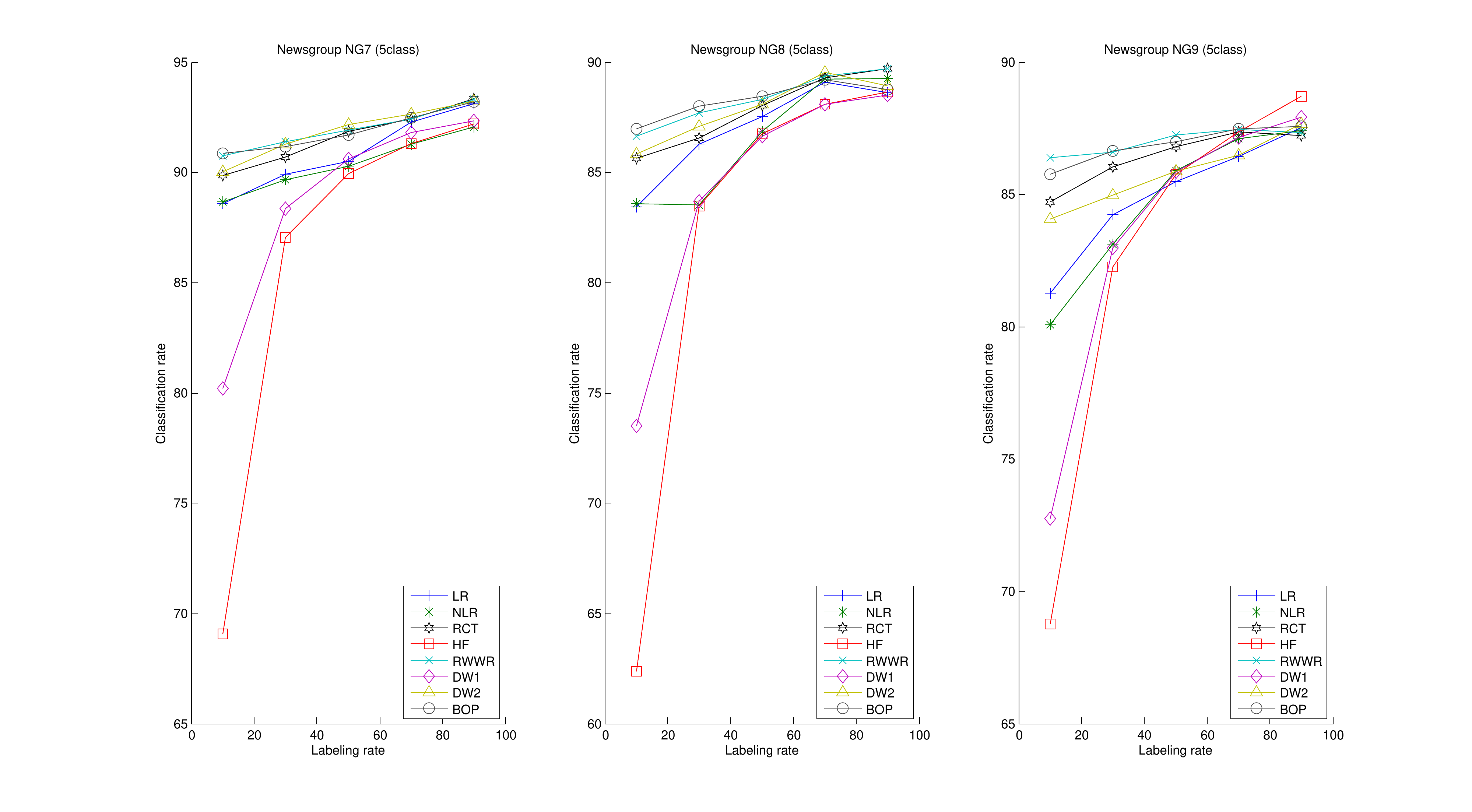}
		\caption{Classification rates in percents, averaged over 20 runs, obtained on partially labeled graphs. Results are reported for the eight methods (RL, RNL, RCT, HF, RWWR, DW1, DW2, BoP) and for five labeling rates (10\%, 30\%, 50\%, 70\%, 90\%). These graphs show the results obtained on the three 5-classes \emph{Newsgroups} data sets.}
		\label{fig: NG5}
		\end{figure*}
		
				\begin{figure*}[t!]
				\centering
		\includegraphics[width=18cm]{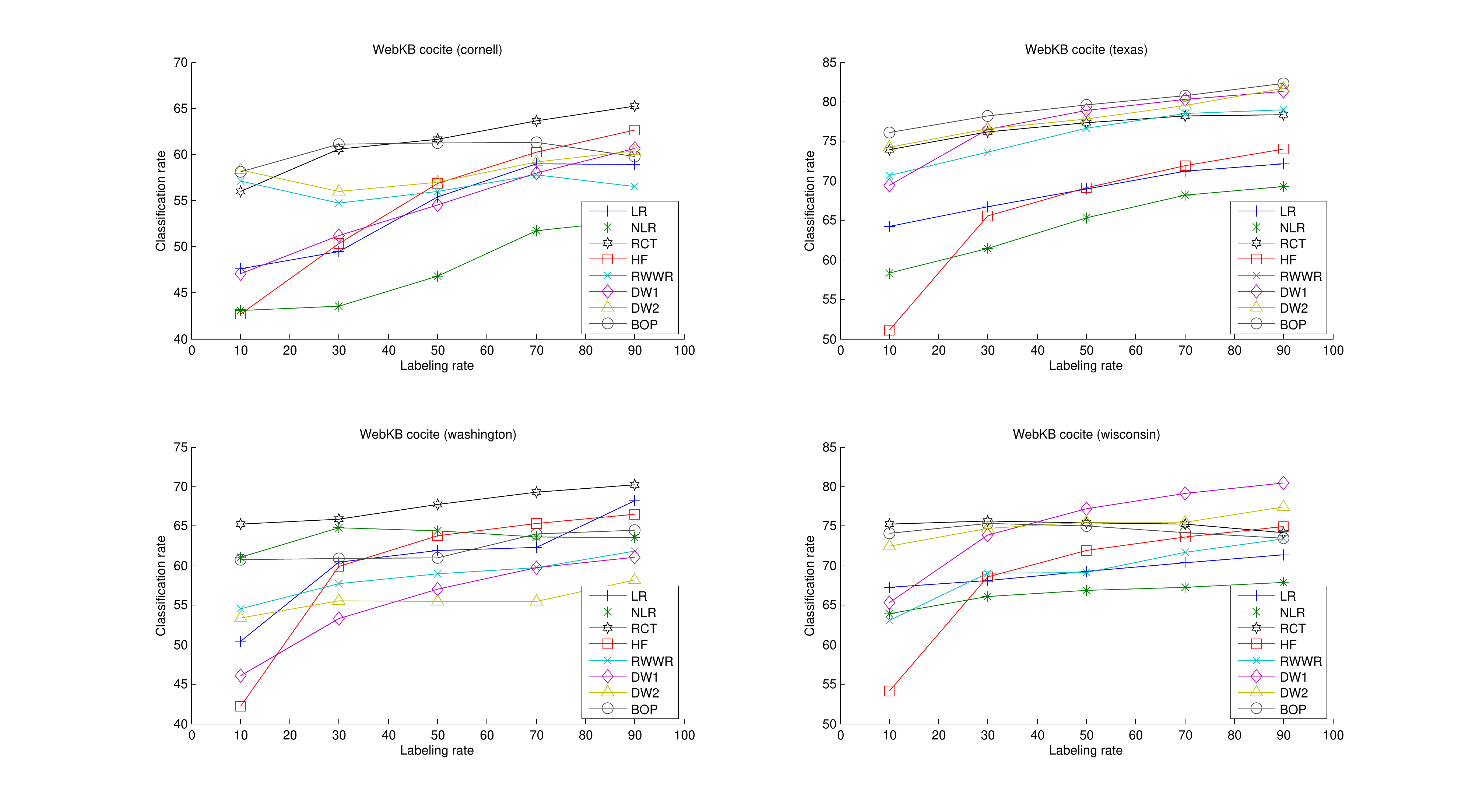}
		\caption{Classification rates in percents, averaged over 20 runs, obtained on partially labeled graphs. Results are reported for the eight methods (RL, RNL, RCT, HF, RWWR, DW1, DW2, BoP) and for five labeling rates (10\%, 30\%, 50\%, 70\%, 90\%). These graphs show the results obtained on the four \emph{WebKB cocite} data sets.}
		\label{fig: Cocite}
		\end{figure*}
		
						\begin{figure}[t!]
	\begin{center}
		\includegraphics[scale = 0.4]{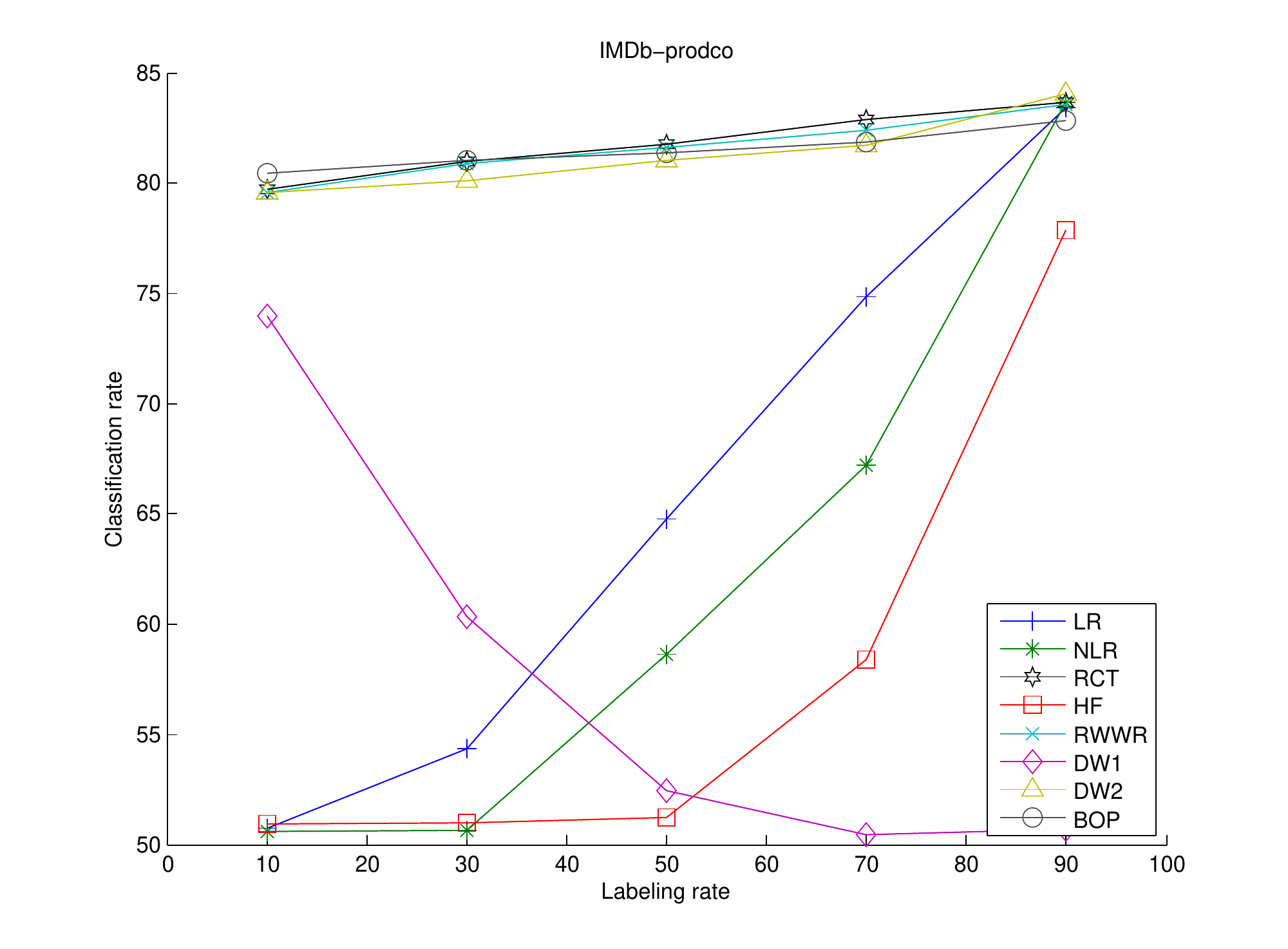}
		\caption{Classification rates in percents, averaged over 20 runs, obtained on partially labeled graphs. Results are reported for the eight methods (RL, RNL, RCT, HF, RWWR, DW1, DW2, BoP) and for five labeling rates (10\%, 30\%, 50\%, 70\%, 90\%). These graphs show the results obtained on the \emph{IMDb-prodco} data set.}
		\label{fig: IMBD}
		\end{center}
		\end{figure}
		

\subsection{Datasets}
\label{DATA}
  The different classifiers are compared on 14 data sets that were used previously for semi-supervised classification: nine \emph{Newsgroups} data sets \cite{Lang95}, the four universities \emph{WebKB cocite} data sets \cite{Macskassy-07}, \cite{Zhou-2005} and the \emph{IMDb prodco} data set \cite{Macskassy-07}.

\emph{Newsgroups}: The \emph{Newsgroups} data set is composed of about 20,000 unstructured documents, taken from 20 discussion groups (newsgroups) of the Usenet diffusion list. 20 Classes (or topics) were originally present in the data set\footnote{The different data sets used for these comparisons are described in Subsection \ref{DATA}. Implementations and datasets are available at http://www.isys.ucl.ac.be/staff/lebichot/research.htm.}. For our experiments, nine subsets related to different topics are extracted from the original data set \cite{Luh-2009}, resulting in a total of nine different data sets. The data sets were built by sampling about 200 documents at random in each topic (three samples of two, three and five classes, thus nine samples in total). Repartition is listed from Table \ref{tab: TabNG}. The extraction process as well as the procedure used for building the graph are detailed in \cite{Luh-2009}.

\emph{WebKB cocite}: These data sets consist of sets of web pages gathered from four computer science departments (four data sets, one for each university), with each page manually labeled into one of six categories: course, department, faculty, project, staff, and student \cite{Macskassy-07}. The pages are linked by co-citation (if $x$ links to $z$ and $y$ links to $z$, then $x$ and $y$ are co-citing $z$), resulting in an undirected graph.
The composition of the data sets is shown in Table \ref{tab: Tabcocite}.\\
\emph{IMDb-prodco}: The collaborative Internet Movie Database (\emph{IMDb}, \cite{Macskassy-07}) has several applications such as making movie recommendations or movie category classification. The classification problem focuses on the prediction of the movie notoriety (whether the movie is a box-office hit or not). It contains a graph of movies linked together whenever they share the same production company. The weight of an edge in the resulting graph is the number of production companies that two movies have in common. The \emph{IMDb-proco} class distribution is shown in Table \ref{tab: TabIMDb}.


\subsection{Experimental methodology}
\label{ExpM}
The classification accuracy will be reported for several labeling rates (10\%, 30\%, 50\%, 70\%, 90\%), i.e. proportions of nodes for which the label is known. The labels of remaining nodes are deleted during the modeling phase and are used as test data during the assessment phase. For each considered labeling rate, 20 random node label deletions were performed (20 runs) and performances are averaged on these 20 runs. For each unlabeled node, the various classifiers predict the most suitable category. Moreover, for each run, a 10-fold nested cross-validation is performed for tuning the parameters of the models. The external folds are obtained by 10 successive rotations of the nodes and the performance of one specific run is the average over these 10 folds. Moreover, for each fold of the external cross-validation, a 10-fold internal cross-validation is performed on the remaining labeled nodes in order to tune the hyper parameters of the classifiers 
(i.e. parameters $\alpha$, $\lambda$ and $\theta$ (see Table \ref{tab:summaryK}) -- methods HF and DW1 do not have any hyper parameter). Thus, for each method and each labeling rate, the mean classification rate averaged on the 20 runs will be reported.

\begin{table*} \centering
\caption{The eight classifiers and the value range tested for tuning their parameters.}
\begin{tabular}{|c|c|c|c|}
\hline
\textbf{Classifier name} & \textbf{Acronym} & \textbf{Parameter} & \textbf{Tested values} \\ \hline
Regularized laplacian kernel                     &RL    & $\lambda > 0$ & $10^{-6},10^{-5},...,10^{6}$ \\ \hline
Regularized normalised laplacian kernel          &RNL   & $\lambda > 0$ & $10^{-6},10^{-5},...,10^{6}$ \\ \hline
Regularized commute-time kernel                  &RCT   & $\alpha \in \left]0,1 \right]$ & $0.1,0.2,...,1$ \\ \hline
Harmonic function                                &HF    & $/ $ & $/$ \\ \hline
Random walk with restart                         &RWWR  & $\alpha \in  \left]0,1 \right]$ & $0.1,0.2,...,1$  \\ \hline
Discriminative random walks                      &DW1   & $/ $ & $/$ \\ \hline
Killing discriminative random walks        &DW2   & $\alpha \in \left]0,1 \right]$ & $0.1,0.2,...,1$ \\ \hline
BoP classifier                                   &BoP   & $\theta >0$ & $10^{-6},10^{-5},...,10^{2}$ \\ \hline 
\end{tabular}
\label{tab:summaryK} 
\end{table*}




\subsection{Results \& discussion}
\label{ResDis}
		
		Comparative results for each method on the fourteen data sets are reported as follows: the results on the nine \emph{NewsGroups} data sets are shown on Fig. \ref{fig: NG2}-\ref{fig: NG5}, the results on the four \emph{WebKB Cocite} data sets are shown on Fig. \ref{fig: Cocite} and the results on the \emph{IMBd-prodco} data set are shown on Fig. \ref{fig: IMBD}.
		
		Statistical significance tests for each labeling rate are detailed from Table \ref{tab: BIGtTEst}. One-side $t$-tests were performed to determine whether or not the performance of a method is significantly superior (p-value lesser than 0.05 on the 20 runs) to another. Table \ref{tab: BIGtTEst} can be read as follows. Each entry indicates on how many data sets (on a total of 14) the row method was significantly better than the column method. At the bottom of each table, the Win/Tie/Lose frequency summarizes how many times the BoP classifier was significantly better (Win), was equivalent (Tie), or was significantly worse (Lose) than each other method.
		
		Moreover, for each labeling rate, the different classifiers have been ordered according to a Borda score ranking. For each data set, each method is granted with a certain number of points, or rating. This number of points is equal to eight if the classifier is the best classifier (i.e., has the best mean classification rate on this data set), seven if the classifier is the second best and so on, so that the worst classifier is granted with only one point. The ratings are then summed across all the considered data sets and the classifiers are sorted by descending total rating. The final ranking, together with the total ratings, are reported from Table \ref{tab: Ranking}.
		
		We observe that the BoP classifier always achieved competitive results since it ranges among the top methods on all data sets. More precisely, the BoP classifier actually tends to be the best algorithm for all labeling rates except for 90\% labeling rate, where it comes third as observed from Table \ref{tab: Ranking} and from Table \ref{tab: BIGtTEst}. The RCT kernel achieves good performance and is the best of the kernel-based classifier (as suggested in \cite{FoussKernelNN-2012}). It is also the best algorithm when the labeling rate is very high (90\%).
		
		Notice that RCT, DW2 and RWWR largely outperform the other algorithms (beside BoP). However, it is difficult to figure out which of those three methods is the best, after BoP. It can be noticed that the DW2 version of the $\mathcal{D}$-walks is more competitive when the labeling rate is low and that it performs much better than the DW1 version, especially for low labeling rates: the Win/Tie/Lose scores for DW2 against DW1 are 7/1/6, 6/1/7, 7/2/7, 13/1/0 and 14/0/0 respectively for 90\%, 70\%, 50\%, 30\%, 10\% percentage of labeling rate.
		
		From the fifth to the eight position, the ranking is less clear since none of the methods is really better than the other. However, all of these methods (NR and RNL as well as HF and DW1) are significantly worse than BoP, RCT, RWWR and DW2. Notice also that the performance of DW1 and HF drops significantly when labeling rate decreases. In addition, the DW1 algorithm provides surprising results on the \emph{IMBd-prodco} data set by raising a classication rate of only 20\%, but this remains anecdotal.

\begin{table*}[htdp]
\caption{One-side $t$-test for all labeling rates. Each entry indicates on how many data sets the row method was significantly better than the column method. On the bottom, the Win/Tie/Lose frequency summarizes how many times the BoP classifier was significantly better (Win), equivalent (Tie) or significantly worse (Lose) than each other method.}
\scriptsize
\begin{center}
\begin{tabular}{llcccccccc}\hline
 & & {\textbf{RL}}& {\textbf{RNL}}& {\textbf{RCT}}& {\textbf{HF}}& {\textbf{RWWR}}& {\textbf{DW1}}& {\textbf{DW2}}& {\textbf{BoP}}\\
\hline
\multirow{9}{0.2cm}{\rotatebox{90}{\textbf{90\% Labelling rate}}}
 & {\textbf{RL}}  & 0  & 9  & 2  & 7   & 4 & 7  & 5  & 4 \\
 & {\textbf{RNL}} & 4  & 0  & 3  & 6   & 4 & 7  & 4  & 4 \\
 & {\textbf{RCT}} & 12 & 10 & 0  & 12  & 3 & 11 & 8  & 9 \\
 & {\textbf{HF}}  & 6  & 7  & 2  & 0   & 4 & 5  & 5  & 4 \\
 & {\textbf{RWWR}}& 10 & 9  & 2  & 12  & 0 & 10 & 8  & 7 \\
 & {\textbf{DW1}} & 5  & 7  & 3  & 2   & 4 & 0  & 6  & 2 \\
 & {\textbf{DW2}} & 9  & 9  & 5  & 9   & 6 & 7  & 0  & 3 \\
 & {\textbf{BoP}} & 7  & 9  & 4  & 8   & 6 & 11 & 6  & 0 \\
\cline{2-10}
 & {\textbf{Win/Tie/Lose BoP}} & 7/3/4  & 9/1/4  & 4/1/9  & 8/2/4 & 6/3/7 & 11/1/2 & 6/5/3 & total: 14\\
\hline
\multirow{9}{0.2cm}{\rotatebox{90}{\textbf{70\% Labelling rate}}}
 & {\textbf{RL}}  & 0  & 9  & 2  & 7   & 4 & 7  & 5  & 4 \\
 & {\textbf{RNL}} & 4  & 0  & 3  & 6   & 4 & 7  & 4  & 4 \\
 & {\textbf{RCT}} & 12 & 10 & 0  & 12  & 3 & 11 & 8  & 9 \\
 & {\textbf{HF}}  & 6  & 7  & 2  & 0   & 4 & 5  & 5  & 4 \\
 & {\textbf{RWWR}}& 10 & 9  & 2  & 12  & 0 & 10 & 8  & 7 \\
 & {\textbf{DW1}} & 5  & 7  & 3  & 2   & 4 & 0  & 6  & 2 \\
 & {\textbf{DW2}} & 9  & 9  & 5  & 9   & 6 & 7  & 0  & 3 \\
 & {\textbf{BoP}} & 7  & 9  & 4  & 8   & 6 & 11 & 6  & 0 \\
\cline{2-10}
 & {\textbf{Win/Tie/Lose BoP}} & 11/3/0  & 11/1/2  & 5/4/5  & 10/2/2 & 7/5/2 & 11/1/2 & 9/1/4 & total: 14\\
\hline
\multirow{9}{0.2cm}{\rotatebox{90}{\textbf{50\% Labelling rate}}}
 & {\textbf{RL}}  & 0  & 9  & 2  & 7   & 4 & 7  & 5  & 4 \\
 & {\textbf{RNL}} & 4  & 0  & 3  & 6   & 4 & 7  & 4  & 4 \\
 & {\textbf{RCT}} & 12 & 10 & 0  & 12  & 3 & 11 & 8  & 9 \\
 & {\textbf{HF}}  & 6  & 7  & 2  & 0   & 4 & 5  & 5  & 4 \\
 & {\textbf{RWWR}}& 10 & 9  & 2  & 12  & 0 & 10 & 8  & 7 \\
 & {\textbf{DW1}} & 5  & 7  & 3  & 2   & 4 & 0  & 6  & 2 \\
 & {\textbf{DW2}} & 9  & 9  & 5  & 9   & 6 & 7  & 0  & 3 \\
 & {\textbf{BoP}} & 7  & 9  & 4  & 8   & 6 & 11 & 6  & 0 \\
\cline{2-10}
 & {\textbf{Win/Tie/Lose BoP}} & 12/1/1  & 12/0/2  & 9/1/4  & 12/0/2 & 9/2/3 & 12/0/2 & 9/2/3 & total: 14\\
\hline
\multirow{9}{0.2cm}{\rotatebox{90}{\textbf{30\% Labelling rate}}}
 & {\textbf{RL}}  & 0  & 9  & 2  & 7   & 4 & 7  & 5  & 4 \\
 & {\textbf{RNL}} & 4  & 0  & 3  & 6   & 4 & 7  & 4  & 4 \\
 & {\textbf{RCT}} & 12 & 10 & 0  & 12  & 3 & 11 & 8  & 9 \\
 & {\textbf{HF}}  & 6  & 7  & 2  & 0   & 4 & 5  & 5  & 4 \\
 & {\textbf{RWWR}}& 10 & 9  & 2  & 12  & 0 & 10 & 8  & 7 \\
 & {\textbf{DW1}} & 5  & 7  & 3  & 2   & 4 & 0  & 6  & 2 \\
 & {\textbf{DW2}} & 9  & 9  & 5  & 9   & 6 & 7  & 0  & 3 \\
 & {\textbf{BoP}} & 7  & 9  & 4  & 8   & 6 & 11 & 6  & 0 \\
\cline{2-10}
 & {\textbf{Win/Tie/Lose BoP}} & 13/1/0  & 12/1/1  & 10/2/2  & 14/0/0 & 9/4/1 & 14/0/0 & 11/1/2 & total: 14\\
\hline
\multirow{9}{0.2cm}{\rotatebox{90}{\textbf{10\% Labelling rate}}}
 & {\textbf{RL}}  & 0  & 9  & 2  & 7   & 4 & 7  & 5  & 4 \\
 & {\textbf{RNL}} & 4  & 0  & 3  & 6   & 4 & 7  & 4  & 4 \\
 & {\textbf{RCT}} & 12 & 10 & 0  & 12  & 3 & 11 & 8  & 9 \\
 & {\textbf{HF}}  & 6  & 7  & 2  & 0   & 4 & 5  & 5  & 4 \\
 & {\textbf{RWWR}}& 10 & 9  & 2  & 12  & 0 & 10 & 8  & 7 \\
 & {\textbf{DW1}} & 5  & 7  & 3  & 2   & 4 & 0  & 6  & 2 \\
 & {\textbf{DW2}} & 9  & 9  & 5  & 9   & 6 & 7  & 0  & 3 \\
 & {\textbf{BoP}} & 7  & 9  & 4  & 8   & 6 & 11 & 6  & 0 \\
\cline{2-10}
 & {\textbf{Win/Tie/Lose BoP}} & 14/0/0  & 13/1/0  & 11/4/2  & 14/0/0 & 9/1/4 & 14/0/0 & 12/1/1 & total: 14\\
\hline
\end{tabular}

\label{tab: BIGtTEst}
\end{center}

\end{table*}


\newcommand\T{\rule{0pt}{2.6ex}}
\newcommand\B{\rule[-1.2ex]{0pt}{0pt}}

\begin{table*}[htdp]
\caption{For each labeling rate, the different classifiers are ranked through a Borda rating (see the text for details). The classifiers are then ranked according to the total rating obtained across all data sets (the larger the better). $l$ stands for labeling rate and the numbers between parentheses are the total ratings.}
\centering
\small
\begin{center}
\begin{tabular}{lllllllll}
\hline
\small{Ranking} & \small{\textbf{First}}& \small{\textbf{Second}}& \small{\textbf{Third}}& \small{\textbf{Fourth}}& \small{\textbf{Fifth}}& \small{\textbf{Sixth}}& \small{\textbf{Seventh}}& \small{\textbf{Last}}\\
\hline
\small{$l$ = 90\%}  & \footnotesize{RCT (86)}  & \footnotesize{RWWR (74)}  &\footnotesize{BoP (71)} & \footnotesize{DW2 (69)} &  \footnotesize{RL (53)} & \footnotesize{HF (53)} & \footnotesize{RNL (50)}  & \footnotesize{DW1 (48)}  \\
\small{$l$ = 70\%}  & \footnotesize{BoP (86)}  & \footnotesize{RCT (82)}   &\footnotesize{RWWR (74)} & \footnotesize{DW2 (65)} &  \footnotesize{HF (59)} & \footnotesize{DW1 (51)} & \footnotesize{RL (44)}  & \footnotesize{RNL (43)}  \\
\small{$l$ = 50\%}  & \footnotesize{BoP (92)}  & \footnotesize{RCT (79)}   &\footnotesize{RWWR (74)} & \footnotesize{DW2 (73)} &  \footnotesize{HF (50)} & \footnotesize{DW1 (50)} & \footnotesize{RNL (46)}  & \footnotesize{RL (40)}  \\
\small{$l$ = 30\%}  & \footnotesize{BoP (104)}  & \footnotesize{DW2 (83)}   &\footnotesize{RWWR (82)} & \footnotesize{RCT (77)} &  \footnotesize{RNL (42)} & \footnotesize{RL (41)} & \footnotesize{HF (41)}  & \footnotesize{HF (34)}  \\
\small{$l$ = 10\%}  & \footnotesize{BoP (103)}  & \footnotesize{RWWR (89)}  & \footnotesize{DW2 (83)} & \footnotesize{RCT (82)} & \footnotesize{RNL (49)} & \footnotesize{RL (46)}  & \footnotesize{HF (35)} & \footnotesize{HF (17)} \\
\hline
\end{tabular}

\label{tab: Ranking}
\end{center}

\end{table*}


\subsection{Computation time}
\label{CTime}

The computational tractability of a method is an important consideration to take into account. Table \ref{tab: running_time} provides a comparison of the running time of all methods. To explore computation time with respect to the number of nodes and the number of classes, the five-classes \emph{Newsgroups} data set number seven (NG7) will be used two times, providing the following variants, NG10 and NG11: 
\begin{itemize}
		\item For NG10, the 499 first nodes are re-labeled class one, and the 499 last nodes are relabeled class two. This provides a two-classes network with 998 nodes.
		\item For NG11, the 100 first nodes are re-labeled class one, the 100 following nodes are re-labeled class two, and so on to get 10 classes (notice that class 9 and 10 have only 99 nodes since NG7 has only 998 nodes). This provides a ten-classes network with 998 nodes.
\end{itemize}
	
	\begin{table*}[htdp]
	\caption{Overview of cpu time in seconds needed to classify all the unlabeled nodes. Results are averaged on 100 runs. The CPU used was an Intel(R)Core(TM)i3 at 2.13 Ghz with 3072 of cache size and 6 GB of RAM and the programming language is Matlab.}
	\centering
\small
\begin{center}
\begin{tabular}{lcccccccc}\hline
 Dataset   & \small{\textbf{RL}}& \small{\textbf{RNL}}& \small{\textbf{RCT}}& \small{\textbf{HF}}& \small{\textbf{RWWR}}& \small{\textbf{DW1}}& \small{\textbf{DW2}}& \small{\textbf{BoP}}\\
\hline
NG1 (2 classes, 400 nodes)  &0.013&0.0433&    0.010&0.012&0.036&0.061&0.064&0.051\\
NG10 (2 classes, 998 nodes) &0.084 &0.422 &0.070&0.109 &0.321 &0.623 &0.639 &0.468\\
NG11 (10 classes, 998 nodes)&0.086&0.445&0.071&0.107&1.167 &2.611&2.683&0.631\\
\\
\small{Ratio NG10/NG1}   &6.28    &9.74    &7.11    &9.07    &8.9  &10.16    &9.98    &9.11\\
\small{Ratio NG11/NG10}  &1.03    &1.06    &1.01    &0.98    &3.63   &4.19    &4.20    &1.35\\
\hline
\end{tabular}

\label{tab: running_time}
\end{center}

\end{table*}


For each method, 100 runs on each of the data sets are performed and the running time is recorded for each run. The 100 running times are averaged and results are reported in Table \ref{tab: running_time}.

We observe that HF is one of the quickest method, but sadly it is not competitive in terms of accuracy, as reported in Subsection (\ref{ResDis}). Notice that two kernel methods, RL and RCT, have more or less the same computation time since the alignment is done in one time for all the classes. RNL, the last kernel method, is slower than RL, HF and RCT. After the HF and the kernel methods, BoP classifier achieves competitive results with the remaining classifiers. The time augmentation when the graph size increases is similar for all methods (except for RL for which the augmentation is smaller), but the BoP classifier has the same advantage than the kernel methods: its computation time does not increase strongly when the number of classes increases. This comes from the algorithm structure: to contrary of RWWR, DW1 and DW2, the BoP classifier does not require a matrix inversion for each class. Furthermore, the matrix inversions (or linear systems of equations to solve) required for the BoP can be computed as far as the graph (through is adjacency matrix) is known, which is not the case with kernel methods. This is a good property for BoP, since it means that rows 1 to 6 of Algorithm 1 can be pre-computed once for all folds in the cross-validation.

\section{Conclusion}
\label{CCL}

This paper investigates an application of the bag-of-paths framework viewing the graph as a virtual bag from which paths are drawn according to a Boltzmann sampling distribution.

In particular, it introduces a novel algorithm for graph-based semi-supervised classification through the bag-of-paths group betweenness, or BoP for short (described in Section \ref{BoPClass}). The algorithm sums the a posteriori probabilities of drawing a path visiting a given node of interest according to a biased sampling distribution, and this sum defines our BoP betweenness measure. The Boltzmann sampling distribution depends on a parameter, $\theta$, gradually biasing the distribution towards shorter paths: when $\theta$ is large, only little exploration is performed and only the shortest paths are considered while when $\theta$ is small (close to $0^{+}$), longer paths are considered and are sampled according to the product of the transition probabilities $p_{ij}^{\text{ref}}$ along the path (a natural random walk).

Experiments on real-world data sets show that the BoP method outperforms the other considered approaches when only a few labeled nodes are available. When more nodes are labeled, the BoP method is still competitive. The computation time of the BoP method is also substantially lower in most of the cases.

Our future work will include several extensions of the proposed approach. Another interesting issue is how to combine the information provided by the graph and the node features in a clever, preferably optimal, way. The interest of including node features should be assessed experimentally. A typical case study could be the labeling of protein-protein interaction networks. The node features could involve gene expression measurements for the corresponding proteins.

Yet another application of the bag-of-paths framework could be the definition of a robustness measure or criticality measure of the nodes. The idea would be to compute the change in reachability between nodes when deleting one node within the BoP framework. Nodes having a large impact on reachability would be then considered as highly critical.

\ifCLASSOPTIONcaptionsoff
  \newpage
\fi



\bibliographystyle{IEEEtran}
\bibliography{Biblio}
\end{document}